\documentclass[10pt,twocolumn,letterpaper]{article}

\usepackage[pagenumbers]{cvpr}

\usepackage{multirow}
\usepackage{float}
\usepackage{dsfont}
\usepackage{makecell}
\usepackage{wrapfig}

\usepackage{xcolor}
\usepackage{colortbl} % 可选，用于给单元格或行上色
\usepackage{amsmath}
\definecolor{ForestGreen}{RGB}{34, 139, 34}
\definecolor{BrickRed}{RGB}{203, 65, 84}
\newcommand{\gain}[1]{\textcolor{ForestGreen}{(+#1\%)}}
\newcommand{\loss}[1]{\textcolor{BrickRed}{(-#1\%)}}

\usepackage{microtype}

\definecolor{cvprblue}{rgb}{0.21,0.49,0.74}
\usepackage[pagebackref,breaklinks,colorlinks,allcolors=cvprblue]{hyperref}

\def\paperID{***} % *** Enter the Paper ID here
\def\confName{CVPR}
\def\confYear{2026}

\title{Focal Guidance: Unlocking Controllability from Semantic-Weak Layers in Video Diffusion Models}

\author{
Yuanyang Yin$^{1,2,3}$\qquad Yufan Deng$^{ 3}$\qquad Shenghai Yuan$^3$ \\ \qquad Kaipeng Zhang$^2$ \qquad Xiao Yang$^3$\qquad Feng Zhao$^1$\footnotemark[1]\qquad\\[3mm]
{$^1$MoE Key Lab of BIPC, USTC\hspace{0.5cm}}
{$^2$Shanghai Innovation Institute\hspace{0.5cm}}
{$^3$ByteDance China\hspace{0.5cm}}\\
}
\begin{document}
\maketitle

\begin{abstract}
The task of Image-to-Video (I2V) generation aims to synthesize a video from a reference image and a text prompt. This requires diffusion models to reconcile high-frequency visual constraints and low-frequency textual guidance during the denoising process. However, while existing I2V models prioritize visual consistency, how to effectively couple this dual guidance to ensure strong adherence to the text prompt remains underexplored.
In this work, we observe that in Diffusion Transformer (DiT)-based I2V models, certain intermediate layers exhibit weak semantic responses (termed Semantic-Weak Layers), as indicated by a measurable drop in text-visual similarity. We attribute this to a phenomenon called Condition Isolation, where attention to visual features becomes partially detached from text guidance and overly relies on learned visual priors.
To address this, we propose Focal Guidance (FG), which enhances the controllability from Semantic-Weak Layers. FG comprises two mechanisms: (1) Fine-grained Semantic Guidance (FSG) leverages CLIP to identify key regions in the reference frame and uses them as anchors to guide Semantic-Weak Layers. (2) Attention Cache transfers attention maps from semantically responsive layers to Semantic-Weak Layers, injecting explicit semantic signals and alleviating their over-reliance on the model's learned visual priors, thereby enhancing adherence to textual instructions.
To further validate our approach and address the lack of evaluation in this direction, we introduce a benchmark for assessing instruction following in I2V models. On this benchmark, Focal Guidance proves its effectiveness and generalizability, raising the total score on Wan2.1-I2V to 0.7250 (+3.97\%) and boosting the MMDiT-based HunyuanVideo-I2V to 0.5571 (+7.44\%).
\end{abstract}
 
\begin{figure*}[thbp]
    \centering
    \includegraphics[width=0.99\textwidth]{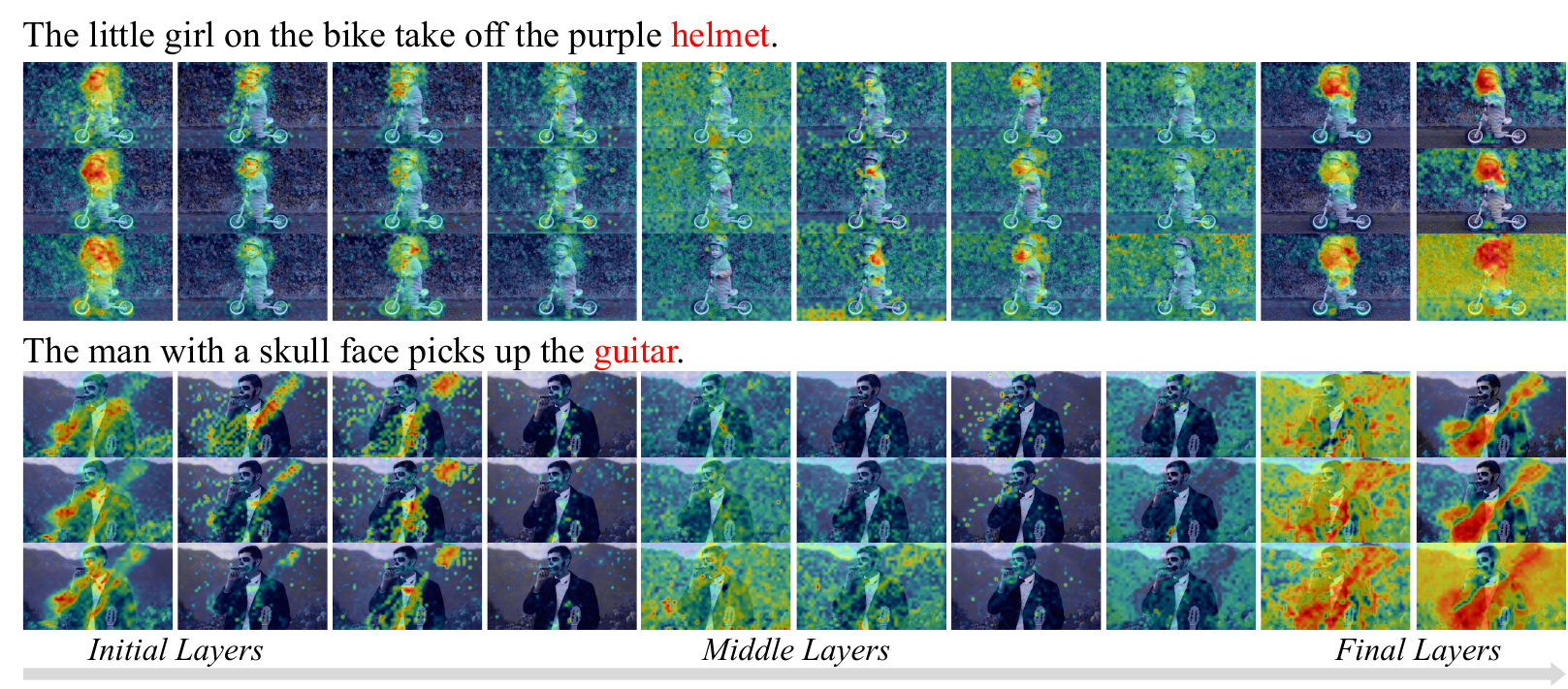}
    \vspace{-2mm}
    \caption{Visualization of semantic alignment within the Wan2.1-I2V \citep{wan2025wan}, quantified by the cosine similarity between visual features and keyword textual features. The features are sampled at evenly spaced inference steps and network layers. The heatmap reveals that the initial and final layers exhibit stronger and more accurate alignment with the target words, while several intermediate layers show noticeably degraded and noisy responses.}
    \label{fig:heatmap}
    \vspace{-4mm}
\end{figure*}

\section{Introduction}\label{sec:intro}
Propelled by diffusion Transformers (DiT)~\citep{ho2020denoising,song2020denoising,peebles2023scalable,ma2024sitexploringflowdiffusionbased,lyu2019advances,sohl2015deep}, the field of Text-to-Video (T2V) generation has achieved remarkable progress ~\citep{blattmann2023align, zhang2023show, bar2024lumiere, tian2021good, jiang2023text2performer, bao2019depth, liu2019deep,ma2025step,hacohen2024ltx,wan2025wan,kong2024hunyuanvideo,peng2025open, yuan2025magictime}. Building upon this, the pursuit of finer-grained controllability has led researchers to extend the paradigm from text-driven synthesis to video generation conditioned on both a starting image and a text prompt, known as the Image-to-Video (I2V) task~\citep{xing2024dynamicrafter,ni2023conditional,zhang2024pia,guo2023animatediff,hu2022make,wan2025wan,kong2024hunyuanvideo,peng2025open}. As a direct extension of the T2V paradigm, I2V task usually incorporates a starting frame as a visual anchor to ensure high fidelity in subject appearance while a text prompt guides the dynamic evolution of the video content. Pioneering works such as WAN~\citep{wan2025wan} and HunyuanVideo~\citep{kong2024hunyuanvideo} have already validated the efficacy of the I2V framework, showcasing its significant potential for producing high-fidelity videos with controllable dynamics.

\begin{figure*}[htbp]
    \centering
    \includegraphics[width=0.99\textwidth]{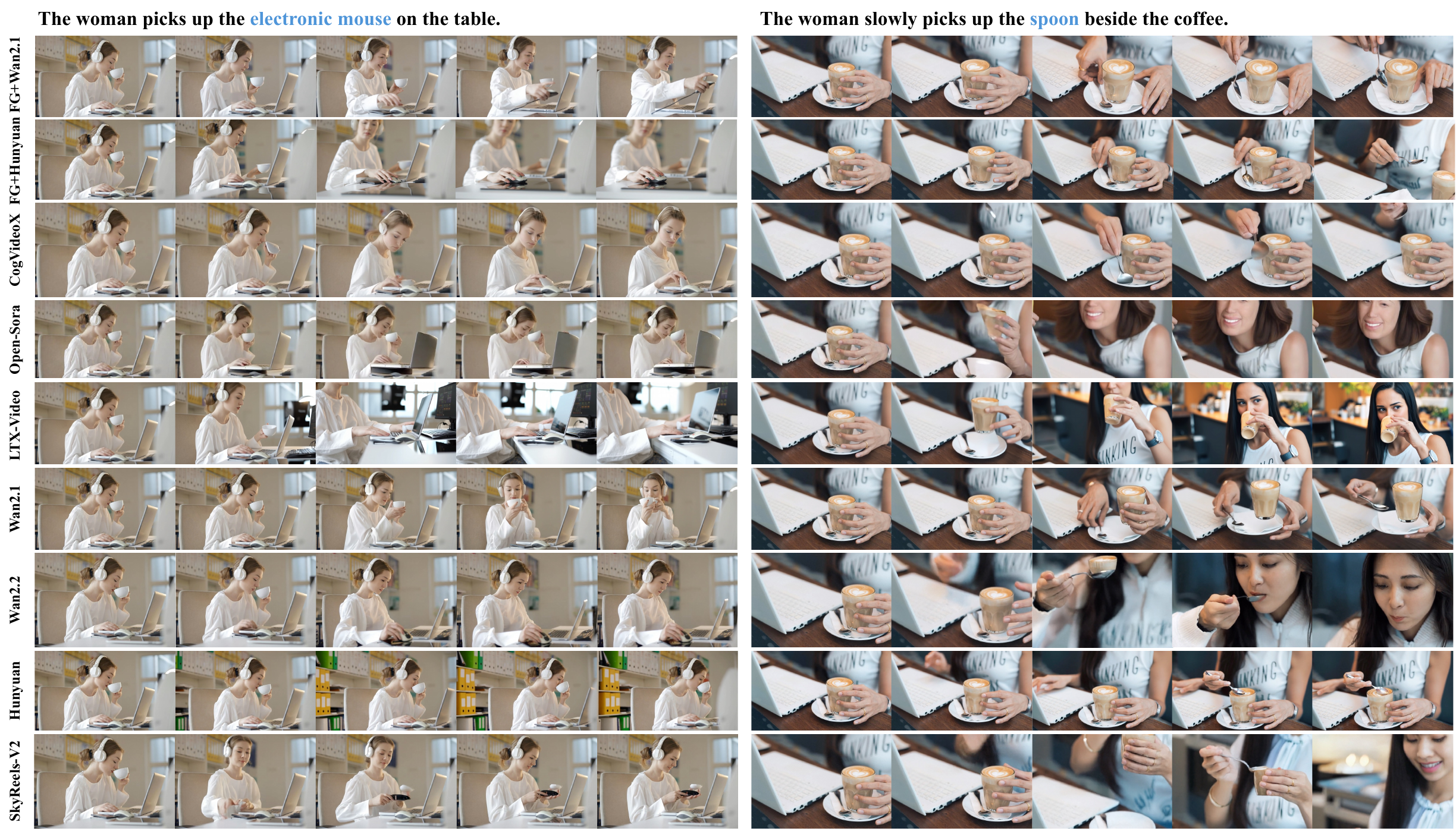}
    \caption{Qualitative comparison of controllability across mainstream open-source I2V models. Existing methods often fail to reliably ground the text instruction in the first-frame reference, leading to instruction non-compliance and hallucinated (or duplicated) visual elements. Our FG approach strengthens text–reference alignment, enabling more accurate instruction following and improved controllability.\textit{\scriptsize (All visual examples in this paper are from public benchmark datasets.)}}
    \label{fig:Qualitative}
    \vspace{-4mm}
\end{figure*}

Despite its promise, a central challenge in Image-to-Video (I2V) generation lies in harmonizing the conditioning signals from the initial frame and the text prompt during the denoising process. Ideally, the model must preserve high-frequency visual details (e.g. subject identity, texture, and style) from the reference image while faithfully executing the motion and semantic transformations dictated by the text. However, even the state-of-the-art I2V models~\citep{wan2025wan,kong2024hunyuanvideo,peng2025open,yang2024cogvideox} struggle to maintain this delicate balance, frequently prioritizing the visual condition and internal priors over textual directives (as shown in~\Cref{fig:Qualitative}). 

Current I2V research, however, has predominantly focused on enhancing temporal consistency and aesthetic quality, leaving the fundamental issue of prompt adherence relatively under-explored. Attempts to address this problem are often indirect and limited to the training phase. For instance, some methods initialize I2V models with weights from a T2V model, hoping to inherit its strong text-responsiveness~\citep{wan2025wan,kong2024hunyuanvideo,peng2025open,yang2024cogvideox}. Others employ techniques like crafting prompts that first describe the reference image in detail to encourage alignment between the two modalities~\citep{chen2025skyreels}.  
Recent interpretability studies in Transformer and DiT-based generation models have shown that semantic responsiveness often varies across layers and that text-based conditioning signals can be partially overshadowed by visual priors~\citep{chefer2021transformer,bousselham2025legrad,tinaz2025emergence,zeng2024decoding}. Therefore, a principled understanding of the underlying causes of prompt mis-alignment becomes essential. To move beyond current limitations in controllability and unlock the true potential of I2V, it is crucial to first diagnose and then rectify the underlying causes of this phenomenon.

\begin{figure}[t!]
    \centering
    \begin{subfigure}[t]{0.49\textwidth}
        \centering
        \includegraphics[width=\textwidth]{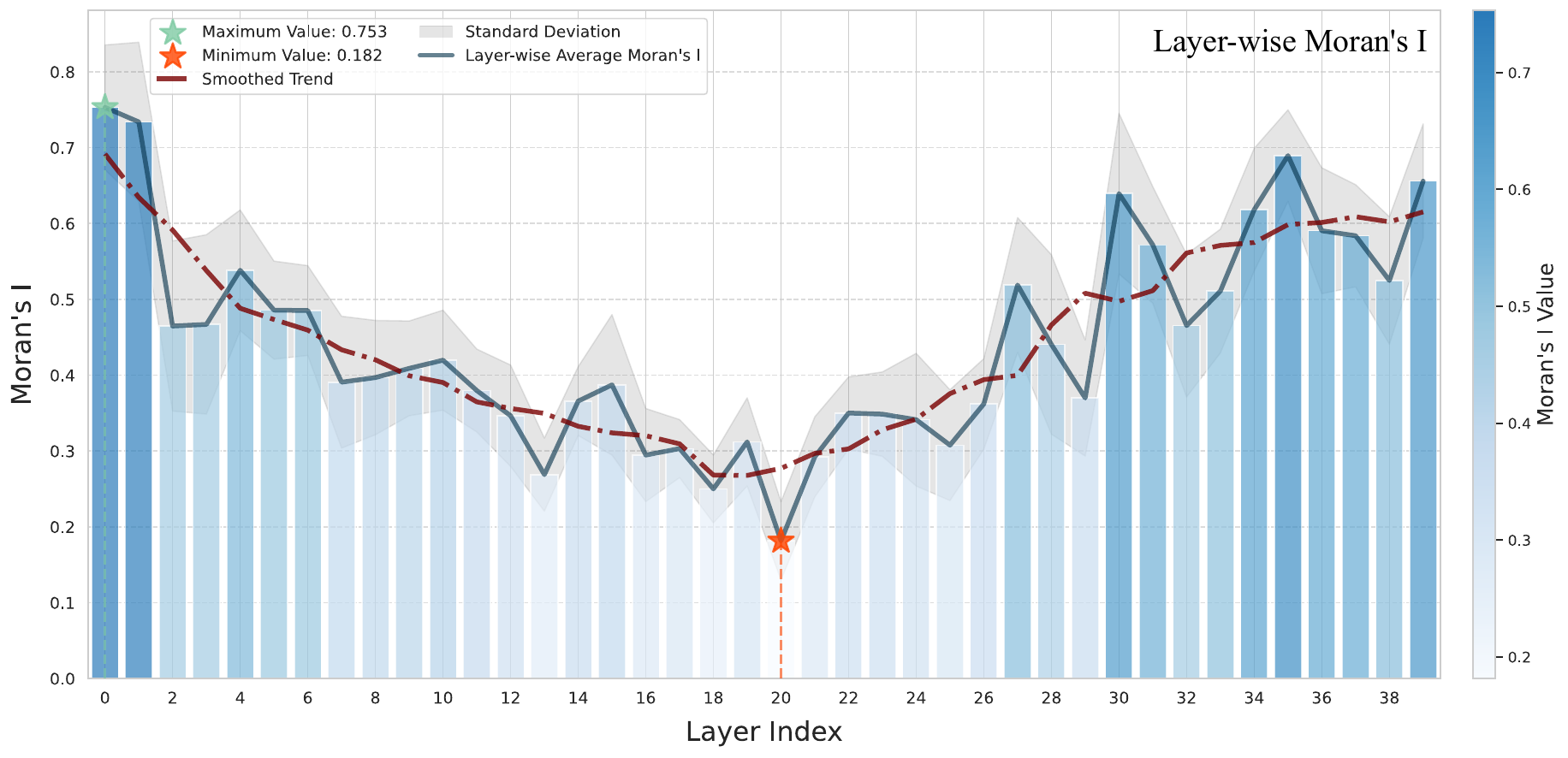}
        \caption{Moran’s I of similarity between visual and textual features.}
        \label{fig:MoranI}
    \end{subfigure}
    \hfill
    \begin{subfigure}[t]{0.477\textwidth}
        \centering
        \includegraphics[width=\textwidth]{fig/Std.png}
        \caption{Standard deviation of similarity between visual and textual features.}
        \label{fig:Std}
    \end{subfigure}
    \caption{Statistical analysis of visual-textual similarity across 50 samples. We evaluate the semantic responsiveness of DiT layers by measuring Moran’s I (~\Cref{fig:MoranI}) and standard deviation (~\Cref{fig:Std}) of normalized visual-textual similarity maps. Consistent with the results in Fig.~\ref{fig:heatmap}, the initial and final layers show stronger and more stable responses to textual keywords, while intermediate layers exhibit weakened semantic alignment.}
    \label{fig:findings}
    \vspace{-4mm}
\end{figure}

Our investigation reveals that poor prompt adherence in DiT-based I2V models originates from the emergence of \textbf{Semantic-Weak Layers} where Moran’s I of text–visual similarity sharply declines from 0.76 to 0.19, indicating a collapse in semantic alignment (see \Cref{fig:findings} and \Cref{fig:heatmap}). These layers undermine text-driven guidance during denoising, ultimately impairing the model’s ability to follow semantic instructions. A key factor that exacerbates this phenomenon is \textbf{Condition Isolation} where the three primary conditioning signals (the VAE-encoded reference frame, image encoder features, and text embeddings) are injected into the model in a relatively isolated manner. This lack of fine-grained alignment increases the likelihood that specific layers fail to establish precise correspondence between textual concepts and their visual counterparts in the initial frame, thereby reinforcing the tendency toward Semantic-Weak Layers and weakening prompt adherence.

Based on these findings, we propose \textbf{Focal Guidance}, a lightweight and principled framework that unlocks semantic controllability in DiT-based I2V models. It consists of two complementary mechanisms (shown in~\Cref{fig:Method} ).
\textbf{Fine-grained Semantic Guidance (FSG)} mitigates conditioning isolation by explicitly aligning textual keywords with their corresponding visual regions in the reference frame, enhancing cross-modal consistency.
\textbf{Attention Cache (AC)} transfers structured attention patterns from semantically responsive layers to weaker ones, reinforcing textual guidance where it tends to collapse.
Together, these mechanisms reestablish coherent text–visual correspondence across layers, significantly improving prompt adherence.
To facilitate rigorous evaluation, we further introduce a benchmark dedicated to assessing instruction-following in I2V models.
Our contributions are summarized as follows:
\begin{itemize}
\item We identify Condition Isolation as the root cause of Semantic-Weak Layers, which in turn leads to poor prompt adherence in DiT-based I2V models, providing a foundation for understanding controllability loss.
\item We propose Focal Guidance, a lightweight framework that directly addresses these issues through Fine-grained Semantic Guidance and the Attention Cache, enabling fine-grained semantic control.
\item We introduce a new benchmark for evaluating instruction-following in I2V models. On this benchmark, Focal Guidance boosts the performance of leading open-source models: improving Wan2.1-I2V by \textbf{+3.97\%} and the MMDiT-based HunyuanVideo-I2V by \textbf{+7.44\%}.
\end{itemize}

\section{Related Work}
\footnotetext{\scriptsize All visual examples are from public benchmark datasets.}
\paragraph{Image-to-Video Generation Models}
Image-to-Video (I2V) generation aims to synthesize a dynamic video sequence from a single static image, enabling richer digital content creation and visual storytelling. Early I2V methods primarily relied on motion priors, reference videos, or modeling specific physical phenomena such as fluids and hair~\citep{siarohin2021motion, chuang2005animating, cheng2020time, shalev2022image, siarohin2019first, zhao2022thin, zhao2021sparse, wang2022latent, mahapatra2022controllable, okabe2009animating, xiao2023automatic}, which constrained their generality and flexibility. With the emergence of U-Net-based diffusion models, approaches such as~\citep{girdhar2024emuvideofactorizingtexttovideo, chen2023seine, lei2024animateanything} introduced conditional control over the first frame by fusing the input image features with noise, while subsequent works like~\citep{xing2024dynamicrafter, zhang2024moonshot} further improved conditioning through cross-attention layers, enhancing video fidelity and consistency. Leveraging advances from Text-to-Video (T2V) generation with DiT-based architectures~\citep{lu2023vdt, ma2024latte, gao2024lumina}, contemporary I2V methods~\citep{xing2024dynamicrafter, ni2023conditional, zhang2024pia, guo2023animatediff, hu2022make, wan2025wan, kong2024hunyuanvideo, peng2025open} build upon pre-trained T2V models to achieve higher visual fidelity, controllability, and temporal coherence, demonstrating the strong potential of the I2V paradigm.
\vspace{-2mm}
\paragraph{Controllable Video Generation}
Controllable video generation methods exploit explicit guidance through different signals: bounding boxes to guide object motion and appearance~\citep{huang2025fine, li2025trackdiffusion, ma2024trailblazer, wang2024boximator, wu2024motionbooth, deng2025magref,yuan2025identity}, trajectories for specific paths~\citep{qiu2024freetraj, shi2024motion, wu2024draganything}, or 3D camera parameters for perspective control~\citep{yu2024viewcrafter, wang2024motionctrl, wang2025akira, hou2024training}. While effective, these approaches require precise external signals and labeled data, leaving intrinsic controllability of base I2V models underexplored.
\vspace{-2mm}
% \paragraph{Interpretability and Conditioning in Generative Models}
% Recent work has examined internal semantics and conditioning mechanisms in generative models.~\citep{chefer2021transformer} found that Transformer layers exhibit semantic responsiveness.~\citep{bousselham2025legrad} found non-monotonic feature formation sensitivity in ViTs, with middle layers showing the lowest semantic selectivity. In diffusion models, Kim et al.~\citep{tinaz2025emergence} reported that U-Net mid-blocks have weaker semantic expression but stronger stylistic control, while~\citep{zeng2024decoding} identified latent space biases where textual signals can be overshadowed by visual priors, causing condition detachment. In DiT-based generation models, semantic representations often emerge later than geometric features~\citep{velez2025image}, and~\citep{helbling2025conceptattention} observed non-uniform text–visual similarity across DiT layers. Attention-intervention methods like ~\citep{chefer2023attend} aim to re-weight attention to enhance semantic guidance. Building on these findings, our work systematically quantify Semantic-Weak Layers in DiT-based I2V models, trace their origin to \textit{Condition Isolation}, and introduce Focal Guidance to restore controllability via fine-grained semantic intervention.
\paragraph{Interpretability and Conditioning in Generative Models}
Interpretability research reveals inconsistent conditioning in generative models. Across architectures like Transformers and ViTs, semantic responsiveness is non-uniform, with middle layers often being the least selective~\citep{chefer2021transformer, bousselham2025legrad}. Similarly, in diffusion models, U-Net mid-blocks exhibit weaker semantic expression~\citep{tinaz2025emergence}, and textual signals can be overshadowed by visual priors, causing "condition detachment"~\citep{zeng2024decoding}. This issue persists in DiT-based models, which show non-uniform text-visual similarity and delayed semantic emergence~\citep{helbling2025conceptattention, velez2025image}. While existing attention interventions offer general correctives~\citep{chefer2023attend}, they do not address a specific root cause. While these disparate findings hint at a common problem, our work provides the first systematic diagnosis in the I2V domain. We identify "Semantic-Weak Layers," trace their origin to a mechanism we term \textbf{\textit{Condition Isolation}}, and introduce Focal Guidance, a targeted intervention built on this diagnosis to restore controllability.

\section{Background and Problem Formulation}
In this section we introduce the fundamental principles of diffusion models then analyze two key issues, Semantic-weak Layers and Conditioning Isolation in I2V models, which directly affect the controllability.

\subsection{Rectified Flow for Video Generation}
Recent state-of-the-art video generation models have increasingly adopted Rectified Flow~\citep{liu2022flow, esser2024scalingrectifiedflowtransformers}, an Ordinary Differential Equation (ODE) based generative framework known for its efficient sampling and straight training paths.
Given a video \( v \in \mathbb{R}^{F \times H \times W \times 3} \), it is first encoded by a VAE encoder \( E \) into a latent representation \( z_0 = E(v) \in \mathbb{R}^{F' \times H' \times W' \times C} \), where the spatial dimensions are typically downsampled.
Rectified Flow defines a linear interpolation path between the data latent \( z_0 \) and a standard Gaussian noise sample \( z_1 \sim \mathcal{N}(0, I) \). For any time \( t \in [0, 1] \), an intermediate latent \( z_t \) on this path is given by:
\begin{equation}
    z_t = (1-t)z_1 + t z_0.
\end{equation}
The model is trained to predict the velocity field along this path. The objective is to minimize the difference between the predicted velocity \( v_{\theta}(z_t, t, c) \) and the path's ground-truth constant velocity, which is \( (z_0 - z_1) \):
\begin{equation}
\mathcal{L} = \mathbb{E}_{z_0, z_1, c, t} \left[ \| v_{\theta}(z_t, t, c) - (z_0 - z_1) \|_2^2 \right],
\label{eq:loss}
\end{equation}
where \( c \) represents the conditioning information (e.g., text and image embeddings). During inference, one starts with a random noise sample \( z_1 \sim \mathcal{N}(0, I) \) and integrates the learned velocity field \( v_{\theta} \) from \( t=1 \) to \( t=0 \) using a numerical ODE solver to obtain the final data latent \( z_0' \).

\subsection{Issues in Current DiT-based I2V Models}
Current DiT-based Image-to-Video (I2V) models have demonstrated remarkable progress in generating videos. Nevertheless, they still face fundamental limitations that hinder their controllability, particularly in integrating the initial image and text prompt. At the core of these limitations lies the emergence of \textbf{Semantic-Weak Layers}, which over-rely on DiT's internal priors and weaken the textual influence. A key structural reason behind this phenomenon is \textbf{Conditioning Isolation}, which restricts the interaction between visual and textual conditions. Together, Conditioning Isolation increases the likelihood of Semantic-Weak Layers, thereby reducing the I2V model’s ability to generate videos that are both visually consistent and text-faithful.
\vspace{-2mm}
\paragraph{Conditioning Isolation} 
One major structural factor contributing to the emergence of the Semantic-Weak layers is the relatively independent injection of multiple conditioning signals, namely the VAE-encoded reference image $z_{\text{ref}} \in \mathbb{R}^{F' \times H' \times W' \times C}$, visual condition features $\mathbf{c}_{\text{img}} \in \mathbb{R}^{N \times D_v}$ extracted by an image encoder, and textual condition features $\mathbf{c}_{\text{text}} \in \mathbb{R}^{M \times D_t}$ obtained from a text encoder. In cross-attention–based architectures~\citep{wan2025wan}, $z_{\text{ref}}$ is concatenated with the first-frame noise latent along the channel dimension, while $\mathbf{c}_{\text{text}}$ and $\mathbf{c}_{\text{img}}$ are injected via cross-attention mechanism. In MMDiT-style designs~\citep{kong2024hunyuanvideo,yang2024cogvideox}, all condition tokens are concatenated along the token dimension before attention.  
Although these designs permit interaction within the Transformer layers, the three modalities originate from \emph{heterogeneous representation spaces}: $z_{\text{ref}}$ encodes high-frequency spatial details, $\mathbf{c}_{\text{text}}$ provides low-frequency semantic guidance, and $\mathbf{c}_{\text{img}}$ captures mid-level visual semantics. Without explicit pre-alignment, the model must learn spatial–semantic correspondences solely through generic attention weights, which is inherently difficult. As a result, semantic entities in $\mathbf{c}_{\text{text}}$ often fail to align with their spatial counterparts in $z_{\text{ref}}$, producing weak grounding at the initial frame (as shown in~\Cref{fig:heatmap}). This weak grounding propagates through temporal denoising, creating fertile ground for semantic-weak layers to emerge.
\vspace{-2mm}
\paragraph{Semantic-Weak Layers} 
As a direct consequence, current DiT-based I2V models exhibit semantic-weak layers which respond weakly to the text prompt. This weak semantic responsiveness suggests that the model, lacking strong textual guidance, may consequently default to its learned internal priors (i.e., generic motion patterns and stylistic biases learned from the large-scale pre-training dataset, rather than the specific instructions in the prompt). This weak semantic responsiveness reduces the constraint of textual instructions during denoising, leading to a misalignment between the intended text-driven transformations and the generated video content as shown in~\Cref{fig:Case}.

To quantify semantic responsiveness, we evaluate each layer's attention to the text prompt using two complementary metrics: \textbf{Moran's I}~\citep{david2023image} and \textbf{Standard Deviation} of the normalized similarity maps between visual and textual features. Moran’s I measures the spatial autocorrelation indicates whether the latent representation at a given layer exhibits a clear and spatially coherent response to the text feature. Let $\mathbf{A}_l \in \mathbb{R}^{F' \times H' \times W'}$ denote the normalized similarity map between the visual features $z_t^l$ and the text features $c_\text{text}$ at layer $l$. For each frame $f \in \{1,\dots,F'\}$, we extract its 2D similarity map $\mathbf{A}_l^{(f)} \in \mathbb{R}^{H' \times W'}$ and flatten it into $\{x_i^{(f)}\}_{i=1}^{H'W'}$. The Moran's I for a given frame $f$ is computed as:
\begin{equation}
I_l^{(f)} = \frac{H'W' \sum_{i,j} w_{ij} (x_i^{(f)} - \bar{x}^{(f)})(x_j^{(f)} - \bar{x}^{(f)})}{\sum_{i} (x_i^{(f)} - \bar{x}^{(f)})^2},
\end{equation}
where $\bar{x}^{(f)}$ is the mean attention value within frame $f$, and $w_{ij}$ is an element of a spatial weight matrix, where $w_{ij}=1$ if pixels $i$ and $j$ are adjacent (using 8-connectivity), and $w_{ij}=0$ otherwise. The layer-wise Moran’s I is then obtained by averaging across all frames:
\begin{equation}
I_l = \frac{1}{F'} \sum_{f=1}^{F'} I_l^{(f)}.
\end{equation}
We then define the layers as semantic-weak layers based on their lower Moran's I values.

As a complementary measure, we use the standard deviation of the normalized similarity to assess the distinctiveness of the text-conditioned attention patterns. A higher standard deviation indicates a more focused and less uniform attention pattern, reflecting a more pronounced semantic significance. For each layer, we compute its score by averaging the standard deviations across all frames:
\begin{equation}
    \mathrm{Std}_l = \frac{1}{F'} \sum_{f=1}^{F'} \sigma(\mathbf{A}_l^{(f)}),
\end{equation}
where $\sigma(\cdot)$ denotes the standard deviation operator.

\begin{figure*}[htbp]
    \centering
    \includegraphics[width=0.99\textwidth]{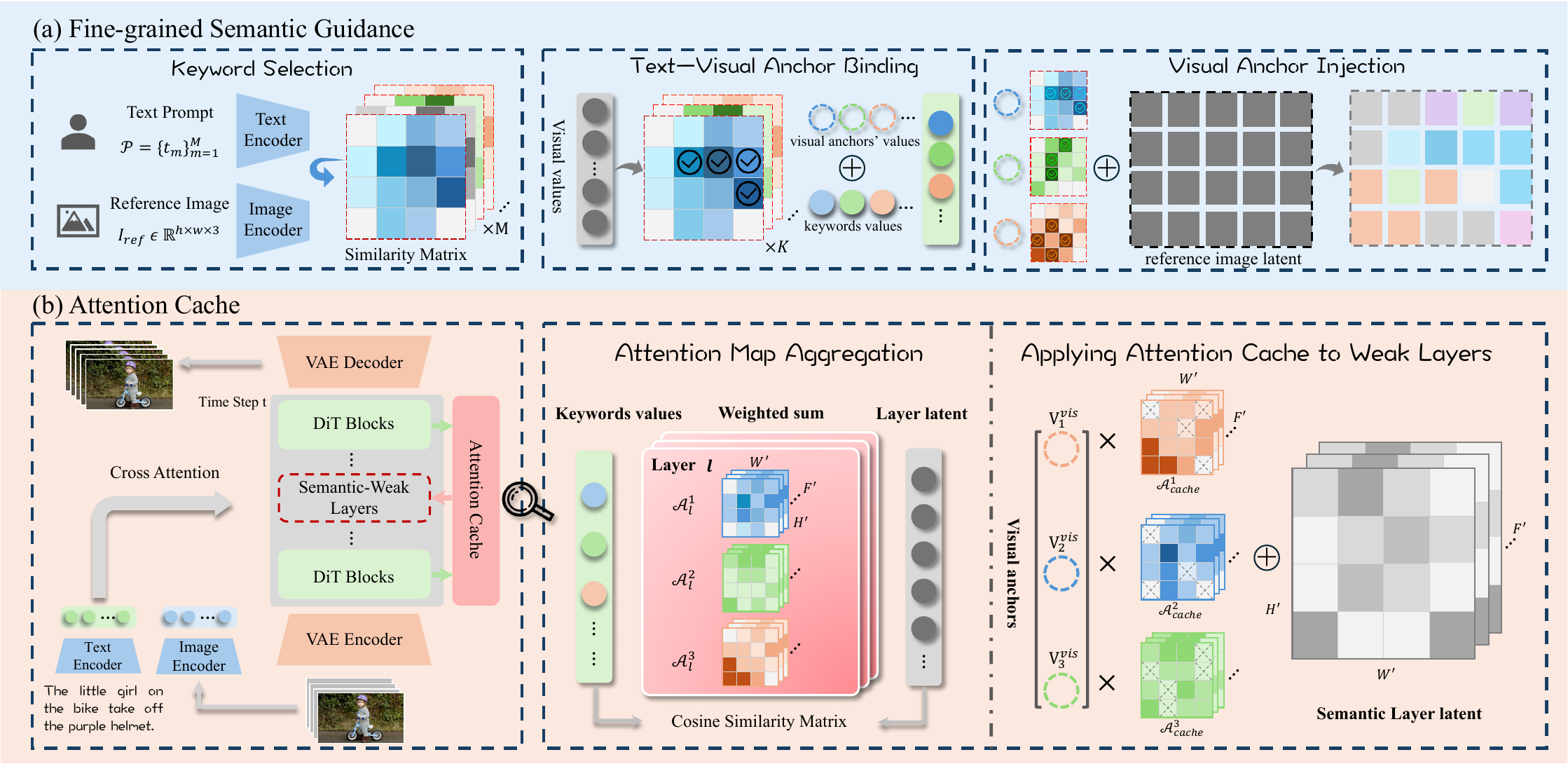}
    \caption{Overview of the Focal Guidance framework. FG consists of two main components: Fine-grained Semantic Guidance and Attention Cache. (a) Fine-grained Semantic Guidance enhances the accuracy of information conditioning and reduces the model's learning complexity by coupling the fine-grained relationships among the VAE-encoded reference frame, image encoder features, and text conditions. (b) Attention Cache leverages the semantic-responsive layers' attention patterns to guide the injection of conditions into layers with weak semantic responses.}
    \label{fig:Method}
    \vspace{-4mm}
\end{figure*}

\section{Method: Focal Guidance Framework}
This section presents Focal Guidance, a framework addressing controllability failures stemming from Semantic-Weak Layers in DiT-based I2V models via two mechanisms: Fine-grained Semantic Guidance (FSG), which couples multi-modal conditions to reduce Conditioning Isolation, and Attention Cache, which transfers structured semantic attention from strong to weak layers to enhance semantic guidance.

\subsection{Fine-grained Semantic Guidance (FSG)}
FSG is designed to alleviate the \emph{conditioning isolation} observed in Semantic-Weak Layers by explicitly coupling textual concepts with their corresponding visual regions in the reference frame. Unlike conventional approaches that rely solely on the Transformer to learn these associations implicitly, FSG injects \emph{visual anchors} into both the text and visual features, thereby establishing a fine-grained cross-modal correspondence before attention computation as shown in~\Cref{fig:Method}(a).
\vspace{-1mm}
\paragraph{Keyword Selection via Text--Image Similarity}
Given a text prompt $\mathcal{P}$ and a reference image $I_{\text{ref}}$, we first employ the visual encoder in the I2V model $\Phi_{\text{img}}(\cdot)$ to extract spatial visual tokens from the second-to-last layer, denoted as $c_{\text{img}}=\{\mathbf{v}_n\}_{n=1}^N \in \mathbb{R}^{D_v}$. Since the visual encoder is aligned with text space (e.g. CLIP), we then use the associated text encoder $\Phi_{\text{text}}(\cdot)$, to convert the prompt $\mathcal{P}$ into text tokens $\{\mathbf{t}_m\}_{m=1}^M \in \mathbb{R}^{D_v}$. For each text token $\mathbf{t}_m$, we compute its negative cosine similarity~\citep{li2023clip} with every spatial position in $c_{\text{img}}$:
\begin{equation}
S_{m,n} = -\frac{\mathbf{t}_m^\top \mathbf{v}_n}{\|\mathbf{t}_m\|_2 \, \|\mathbf{v}_n\|_2},
\label{eq:clip_sim}
\end{equation}
where $\mathbf{v}_n$ denotes the visual token at spatial position $n$. Text words are selected into the keyword set $\mathcal{K}$ if their maximum similarity $\max_n S_{m,n}$ exceeds a predefined threshold $\tau_{\text{sel}}$. 
The visual anchors $V_{\text{anchor}} = \left\{ \sum_{n=1}^N S_{k,n} \cdot \mathbf{v}_n \right\}_{k \in \mathcal{K}}$ are computed as weighted sums of the visual tokens $\mathbf{v}_n$ based on the similarity scores $S_{k,n}$. We then inject the visual anchors into text and the visual features of the Smentic-Weak layers, thereby connecting the isolated conditions.

\vspace{-6mm}
\paragraph{Text–Visual Anchor Binding}
For each selected keyword $k \in \mathcal{K}$ from Eq.~\eqref{eq:clip_sim}, we first project both the textual embedding $\mathbf{t}_k \in \mathbb{R}^{D_t}$ (from the language model e.g. T5) and its corresponding visual anchor $\mathbf{v}_{\text{anchor},k} \in \mathbb{R}^{D_v}$  into the shared DiT latent space:
\begin{equation}
\hat{\mathbf{t}}_k = \mathcal{P}_t(\mathbf{t}_k), \quad 
\hat{\mathbf{v}}_{k,\text{anchor}} = \mathcal{P}_v(\mathbf{v}_{\text{anchor},k}).
\end{equation}
Then  $\hat{\mathbf{t}}_k$ and $\hat{\mathbf{v}}_{\text{anchor}}$ are processed by each layer to produce query ($Q$), key ($K$), and value ($V$) vectors.  
Let $V^{\text{text}}_k = W_v^{\text{text}} \hat{\mathbf{t}}_k$ and $V^{\text{vis}}_k = W_v^{\text{vis}} \hat{\mathbf{v}}_{\text{anchor}}$ denote the value vectors of the text token and the visual anchor, respectively.  
We enhance the text token's value by additive fusion:
\begin{equation}
V^{\text{text}}_k \leftarrow V^{\text{text}}_k + \lambda_{\text{txt}} \cdot V^{\text{vis}}_k,
\label{eq:txt_fuse}
\end{equation}
where $\lambda_{\text{txt}}$ controls the injection strength.  
This design enriches the text token's content representation with spatially grounded visual cues while keeping its query and key vectors unchanged, ensuring the stability of attention patterns.
\vspace{-6mm}
\paragraph{Visual Anchor Injection into Latent Features}
Let $z_{\text{ref}}$ denote the reference frame within the DiT layer's hidden state. 
For each $k \in \mathcal{K}$, its corresponding spatial region $\mathcal{R}_k$ in $z_{\text{ref}}$ is determined based on the similarity map $S_{k,n}$, where a threshold is applied to extract the valid area. 
The visual anchor value representation $\mathbf{V}_k^{\text{vis}}$ is then directly injected into the latent feature map as follows:
\begin{equation}
z_{\text{ref}}^{(u,v)} \leftarrow z_{\text{ref}}^{(u,v)} 
+ \lambda_{\text{lat}} \cdot w_{k}^{(u,v)} \mathbf{V}_k^{\text{vis}},
\quad \forall (u,v) \in \mathcal{R}_k,
\label{eq:latent_inject}
\end{equation}
where $w_{k}^{(u,v)}$ denotes the normalized similarity weight at spatial location $(u,v)$, and $\lambda_{\text{lat}}$ controls the strength of the injection.  This operation plants localized control signals into the generative latent space, ensuring that key objects are preserved and semantically aligned at each step.

\begin{figure*}[!htbp]
    \centering
    \includegraphics[width=0.99\textwidth]{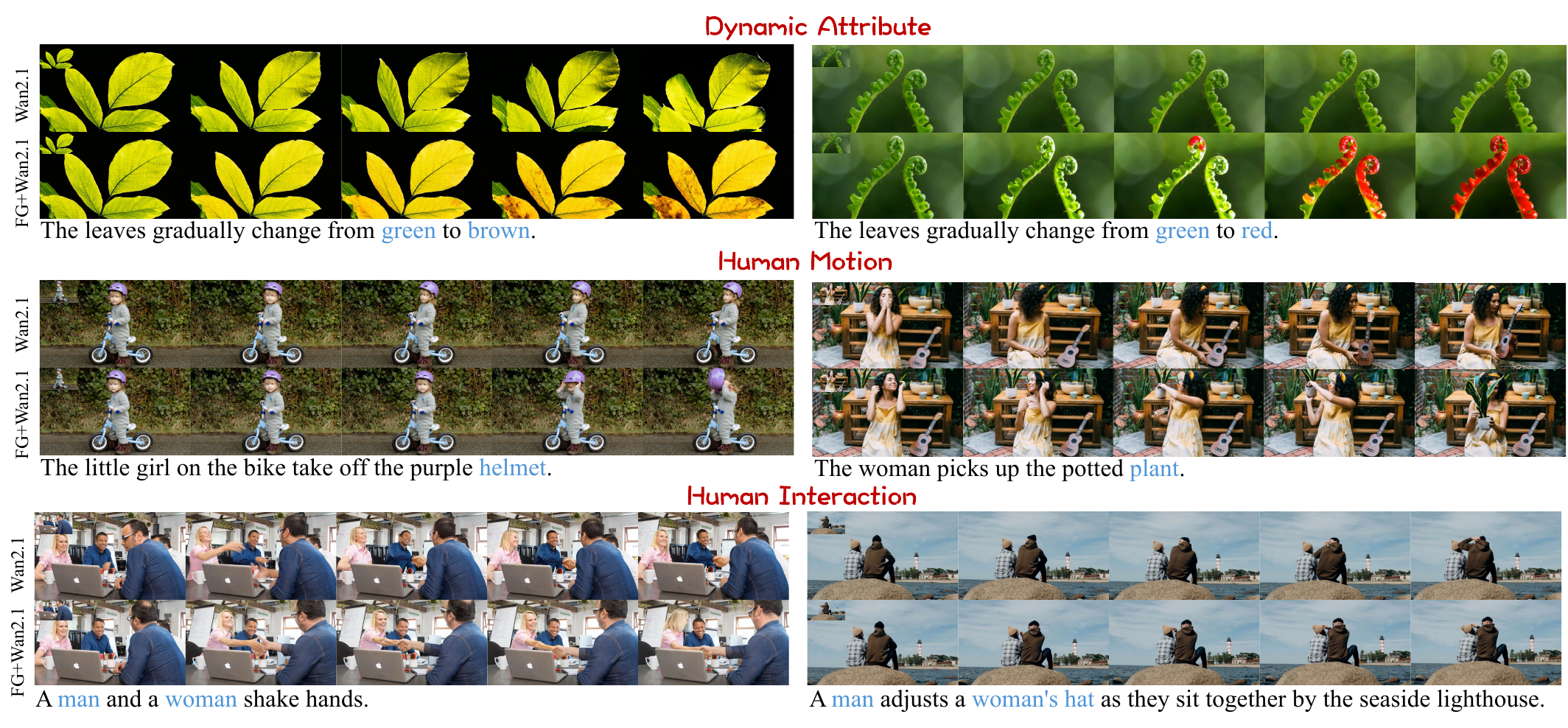}
    \caption{Qualitative ablations on Wan2.1-I2V. We randomly sample cases along three dimensions—Human Motion, Dynamic Attributes, and Human Interaction. With FG, text–reference alignment is strengthened, motions and attributes follow instructions more faithfully.}
    \label{fig:Case}
    \vspace{-4mm}
\end{figure*}
\subsection{Attention Cache}
Fine-grained Semantic Guidance resolves the issue of isolated condition information by establishing a fine-grained binding between the text and the reference frame. This reduces the difficulty of coupling different modal conditions, but the process still relies on the self-modeling capacity of the current layer. To further enhance instruction-following ability, we propose \textbf{Attention Cache mechanism} that reuses attention from semantic-responsive layers to guide the Semantic-Weak layers.

Specifically, the attention cache captures the similarity maps between the text and visual features at the semantic-responsive layers, which record their attention to the text conditioning, and uses it to guide the semantic-weak layers, as shown in ~~\Cref{fig:Method}(b).
\vspace{-6mm}
\paragraph{Attention Aggregation}  
For each layer \( l \) at time step $t$, the similarity map \( \mathcal{A}^l_t \in \mathbb{R}^{K \times F' \times H' \times W'} \) is computed to quantify the cosine similarity between the keyword values \( V(k)_{\text{t}}^l \) and the visual features \( z_t^l \) :
\begin{equation}
    \mathcal{A}^t_{l,k}(u,v) = \frac{V(k)_{\text{t}}^l \, z_t^l(u,v)^\top}{\|V(k)_{\text{t}}^l\|_2 \, \|z_t^l(u,v)\|_2},
\end{equation}
where \( z_t^l(u,v) \) is  the visual features at spatial position \( (u,v) \)and \( V(k)_{\text{t}}^l \) is keyword values. 

We compute a weighted sum of the attention maps across layers to get the attention cache:
\begin{equation}
    \mathcal{A}_{cache}^t = \sum_{l=1}^L \alpha_l \mathcal{A}_l^t,
\end{equation}
where \( \alpha_l \) is the weight for the similarity map at layer \( l \) (set to 0 for semantic-weak layers), and for other layers, \( \alpha_l = \frac{1}{L - m} \), where \( m \) is the number of semantic-weak layers.

\vspace{-2mm}
\paragraph{Applying Attention Cache to Semantic-Weak Layers}  
During both training and inference, the Attention Cache is utilized to guide the attention mechanism in semantic-weak layers. For each Sementic-Weak layer $l_w$, we apply $\mathcal{A}^t_{cache}$ to more accurately inject the text condition into the semantically corresponding regions. This reduces the tendency of these layers to rely solely on the visual priors for denoising, which would otherwise weaken the constraint imposed by the text condition.

Specifically, for each keyword $k \in \mathcal{K}$ (obtained from the FSG procedure~\Cref{eq:clip_sim}), we apply a threshold to $\mathcal{A}^t_{cache,k}$, retaining only the positions with similarity greater than a predefined threshold $\tau_{\text{cache}}$:
\begin{equation}
\mathcal{ A}^t_{cache,k} \leftarrow \mathds{1}_{\{\mathcal{ A}^t_{cache,k} > \tau_{\text{cache}}\}} \cdot \mathcal{A}^t_{cache,k}\ ,
\label{Cache}
\end{equation}
where $\mathds{1}_{\{S_{k,i} > \tau_{\text{cache}}\}}$ is an indicator function that sets all values below the threshold $\tau_{\text{cache}}$ to zero, ensuring that only the most relevant regions are preserved.

In line with the procedure in FSG, instead of directly using the text condition for localization, we employ the visual anchor's value representation $\mathbf{V}_k^{\text{vis}}$ as a guiding reference to assist in the injection of semantic information into Semantic-Weak layers. The visual features $z_{l_w}^t$ of Semantic-Weak layers are updated by $\mathcal{A}^t_{cache}$:
\begin{equation}
z_{l_w}^t \leftarrow z_{l_w}^t + \lambda_{\text{cache}} \cdot \mathcal{A}^t_{cache,k} \cdot \mathbf{V}_k^{\text{vis}}\ ,
\end{equation}
where $k \in \mathcal{K}$ corresponds to the $k$-th keyword.

\section{Experiments}
In this section, we quantitatively assess the effectiveness of Focal Guidance (FG) on two state-of-the-art open-source I2V models: Wan2.1-I2V (CrossDiT-based)\citep{wan2025wan} and HunyuanVideo-I2V (MMDiT-based)\citep{kong2024hunyuanvideo} under small-scale post-training that fine-tunes the Semantic-Weak layers. To address the current gap in evaluation metrics for I2V models, we introduce a new benchmark designed specifically to assess the instruction-following capabilities of image to video generation models. The benchmark evaluates models across three key dimensions: \textbf{dynamic attributes}, \textbf{human motion}, and \textbf{human interaction}.
\subsection{Experimental Setup}
\paragraph{Implementation Details} We utilize an internally video dataset of 12K samples with accurate captions for fine-tuning. FG aims to enhance I2V model controllability with minimal post-training on limited data, and is model-agnostic, applicable to any I2V model. We evaluate FG on the CrossDiT-based Wan2.1-I2V~\citep{wan2025wan} and the MMDiT-based HunyuanVideo-I2V~\citep{kong2024hunyuanvideo}, with full implementation details provided in ~\Cref{appendix:exp}.
\vspace{-4mm}
\paragraph{Metric}
To fill the gap in existing I2V evaluation methods, we propose a comprehensive benchmark assessing controllability across three dimensions: dynamic attributes, human motion, and human interaction. Each dimension is supported by manually annotated datasets and evaluated using a video-based VQA framework~\cite{zheng2025vbench} (see ~\Cref{appendix:metric}). In addition, we adopt Subject Consistency and Background Consistency from the vbench2\_beta\_i2v benchmark~\citep{zheng2025vbench} to measure visual consistency with the reference frame. The final score is computed as the average across all dimensions, providing a holistic measure of model performance.

\begin{table*}[t!]
  \centering
  \resizebox{0.99\textwidth}{!}{%
    % \begin{tabular}{lccccccc}
    %   \toprule
    %   \textbf{Method} & \textbf{Param} & \textbf{I2V Subject} & \textbf{I2V Background} & \textbf{Dynamic Attributes} & \textbf{Human Motion} & \textbf{Human Interaction} & \textbf{Total Score} \\
    %   \midrule
    %   CogVideoX-I2V            & 5B  & 0.9658 & 0.9787 & 0.1279 & 0.6100 & 0.4500 & 0.6265 \\
    %   Open-Sora Plan v1.3      & 2.7B& 0.9630 & 0.9781 & 0.1047 & 0.4300 & 0.4400 & 0.5832 \\
    %   LTX-Video                & 13B & 0.9845 & 0.9893 & 0.2558 & 0.4800 & 0.3500 & 0.6119 \\
    %   Wan2.1-I2V               & 14B & 0.9685 & 0.9870 & \underline{0.3512} & 0.6920 & \underline{0.4880} & \underline{0.6973} \\
    %   Wan2.2-TI2V              & 5B  & 0.9858 & \underline{0.9941} & 0.1512 & 0.7000 & 0.3700 & 0.6402 \\
    %   HunyuanVideo-I2V         & 13B & \textbf{0.9886} & \textbf{0.9942} & 0.1698 & 0.2600 & 0.1800 & 0.5185 \\
    %   SkyReels-V2-I2V          & 14B & \underline{0.9867} & 0.9916 & 0.0465 & \underline{0.7100} & 0.3200 & 0.6110 \\
    %   \bottomrule
    %   FG+Wan2.1-I2V                & 14B & 0.9694 & 0.9875 & \textbf{0.3860} & \textbf{0.7500} & \textbf{0.5320} & \textbf{0.7250} \\
    %   FG+HunyuanVideo-I2V      & 13B & \underline{0.9867}     & 0.9937     & 0.2270     & 0.3480     & 0.2300     & 0.5571 \\
    %   \bottomrule
    % \end{tabular}%
    
    \begin{tabular}{lccccccc}
        \toprule
        \textbf{Method} & \textbf{Param} & \textbf{I2V Subject} & \textbf{I2V Background} & \textbf{Dynamic Attributes} & \textbf{Human Motion} & \textbf{Human Interaction} & \textbf{Total Score} \\
        \midrule
        CogVideoX-I2V            & 5B  & 0.9658 & 0.9787 & 0.1279 & 0.6100 & 0.4500 & 0.6265 \\
        Open-Sora Plan v1.3      & 2.7B& 0.9630 & 0.9781 & 0.1047 & 0.4300 & 0.4400 & 0.5832 \\
        LTX-Video                & 13B & 0.9845 & 0.9893 & 0.2558 & 0.4800 & 0.3500 & 0.6119 \\
        Wan2.1-I2V               & 14B & 0.9685 & 0.9870 & \underline{0.3512} & 0.6920 & \underline{0.4880} & \underline{0.6973} \\
        Wan2.2-TI2V              & 5B  & 0.9858 & \underline{0.9941} & 0.1512 & 0.7000 & 0.3700 & 0.6402 \\
        HunyuanVideo-I2V         & 13B & \textbf{0.9886} & \textbf{0.9942} & 0.1698 & 0.2600 & 0.1800 & 0.5185 \\
        SkyReels-V2-I2V          & 14B & \underline{0.9867} & 0.9916 & 0.0465 & \underline{0.7100} & 0.3200 & 0.6110 \\
        \midrule
        \rowcolor{gray!10} % 给第一行结果加一个浅灰色背景
        FG+Wan2.1-I2V            & 14B & 0.9694 \gain{0.09} & 0.9875 \gain{0.05} & \textbf{0.3860} \gain{9.91} & \textbf{0.7500} \gain{8.38} & \textbf{0.5320} \gain{9.02} & \textbf{0.7250} \gain{3.97} \\
        \rowcolor{gray!10} % 给第二行结果加一个浅灰色背景
        FG+HunyuanVideo-I2V      & 13B & 0.9867 \loss{0.19} & 0.9937 \loss{0.05} & 0.2270 \gain{33.69} & 0.3480 \gain{33.85} & 0.2300 \gain{27.78} & 0.5571 \gain{7.44} \\
        \bottomrule
    \end{tabular}%
  }
  \caption{Quantitative comparison across open-source I2V models. Best scores are in \textbf{bold}; second best are \underline{underlined}. The Total Score is the arithmetic mean of five metrics (I2V Subject, I2V Background, Dynamic Attributes, Human Motion, Human Interaction). Our Fine-grained Guidance (FG) delivers clear controllability gains across two mainstream architectures while preserving subject/background fidelity in image-to-video generation.}

  \label{tab:Main_Results}
\end{table*}

\begin{table*}[!htbp] % 或 table* 如果你在两栏文档并且想跨栏
  \centering
  \resizebox{0.99\textwidth}{!}{%
    % \begin{tabular}{lcccccccc}
    %   \toprule
    %   \textbf{Method} & \textbf{Subject} & \textbf{Background} & \textbf{Motion} & \textbf{Dynamic} & \textbf{Aesthetic} & \textbf{Imaging} & \textbf{I2V} & \textbf{I2V} \\
    %   & \textbf{Consistency} & \textbf{Consistency} & \textbf{Smoothness} & \textbf{Degree} & \textbf{Quality} & \textbf{Quality} & \textbf{Subject} & \textbf{Background} \\
    %   \midrule
    %   Wan2.1-I2V & \underline{0.9375} & 0.9691 & \textbf{0.9765} & 0.5935 & 0.6324 & \textbf{0.7089} & 0.9685 & 0.9870 \\
    %   Wan2.1-I2V w/ post-training & 0.9367 & \textbf{0.9750} & 0.9764 & \textbf{0.6423} & 0.6398 & 0.7067 & \underline{0.9698} & \underline{0.9886} \\
    %   FG+Wan2.1-I2V w/ zero-shot & \textbf{0.9388} & \underline{0.9741} & 0.9764 & 0.5935 & \underline{0.6412} & \underline{0.7088} & \textbf{0.9711} & \textbf{0.9889} \\
    %   FG+Wan2.1-I2V w/ post-training & 0.9372 & 0.9732 & \textbf{0.9765} & \underline{0.6260} & \textbf{0.6432} & 0.7052 & 0.9694 & 0.9875 \\
    %   \bottomrule
    % \end{tabular}%
    \begin{tabular}{lcccccccc}
        \toprule
        \textbf{Method} & \textbf{Subject} & \textbf{Background} & \textbf{Motion} & \textbf{Dynamic} & \textbf{Aesthetic} & \textbf{Imaging} & \textbf{I2V} & \textbf{I2V} \\
        & \textbf{Consistency} & \textbf{Consistency} & \textbf{Smoothness} & \textbf{Degree} & \textbf{Quality} & \textbf{Quality} & \textbf{Subject} & \textbf{Background} \\
        \midrule
        Wan2.1-I2V & 0.9375 & 0.9691 & \textbf{0.9765} & 0.5935 & 0.6324 & \textbf{0.7089} & 0.9685 & 0.9870 \\
        Wan2.1-I2V w/ post-training & 0.9367 \loss{0.09} & \textbf{0.9750} \gain{0.61} & 0.9764 \loss{0.01} & \textbf{0.6423} \gain{8.22} & 0.6398 \gain{1.17} & 0.7067 \loss{0.31} & \underline{0.9698} \gain{0.13} & \underline{0.9886} \gain{0.16} \\
        FG+Wan2.1-I2V w/ zero-shot & \textbf{0.9388} \gain{0.14} & \underline{0.9741} \gain{0.52} & 0.9764 \loss{0.01} & 0.5935\gain{0.00} & \underline{0.6412} \gain{1.39} & \underline{0.7088} \loss{0.01} & \textbf{0.9711} \gain{0.27} & \textbf{0.9889} \gain{0.19} \\
        FG+Wan2.1-I2V w/ post-training & 0.9372 \loss{0.03} & 0.9732 \gain{0.42} & \textbf{0.9765}\gain{0.00} & \underline{0.6260} \gain{5.48} & \textbf{0.6432} \gain{1.71} & 0.7052 \loss{0.52} & 0.9694 \gain{0.09} & 0.9875 \gain{0.05} \\
        \bottomrule
    \end{tabular}%
  } 
  \caption{Impact of post-training data on conventional I2V metrics. Our post-training does not yield noticeable improvements on these metrics, which primarily focus on aesthetics and consistency, while lacking measures of instruction-following ability.}
  \label{tab:vbench}
  \vspace{-4mm}
\end{table*}
\subsection{Main Results}
To ensure fairness, all quantitative results are averaged over five random seeds. We evaluate FG on Wan I2V-14B using the vbench2\_beta\_i2v benchmark~\citep{zheng2025vbench}. As shown in~\Cref{tab:vbench}, we make two key observations: 1) Additional post-training data has little impact on performance, confirming that FG's gains are not due to extra data; 2) Existing I2V metrics, which focus on video quality and consistency, fail to capture the improvements in model responsiveness to textual instructions. As shown in~\Cref{fig:Case}, FG-Wan2.1 demonstrates higher responsiveness than Wan2.1, though this improvement is not reflected in traditional metrics, which emphasize reference-frame fidelity over instruction adherence.

We retain the vbench\_beta\_i2v~\cite{zheng2025vbench} consistency metrics \textbf{I2V\_Subject} and \textbf{I2V\_Background} to measure fidelity to the first-frame reference, and use \textbf{Dynamic Attributes}, \textbf{Human Motion}, and \textbf{Human Interaction} to assess instruction following. The average values across these five dimensions are then considered as the overall score. As shown in \Cref{tab:Main_Results}, Wan2.1-I2V with FG achieves the strongest semantic control, improving the Total Score by \textbf{3.97\%} (0.6973$\rightarrow$0.7250). FG is also effective on the MMDiT-based HunyuanVideo-I2V, where combining FG raises the Total Score by \textbf{7.44\%} (0.5185$\rightarrow$0.5571).
% \begin{table}[h!]
%   \centering
%   \resizebox{0.48\textwidth}{!}{%
%     \begin{tabular}{lccc}
%       \toprule
%       \textbf{Method} & \textbf{Dynamic Attributes} & \textbf{Human Motion} & \textbf{Human Interaction} \\
%       \midrule
%       Wan2.1 & 0.3512 & 0.6920 & 0.4880 \\
%       Wan2.1 w/ post-training & 0.3628 & 0.6980 & 0.5140 \\
%       FG+Wan2.1 w/ zero-shot & 0.3512 & 0.7020 & 0.5220 \\
%       Wan2.1 w/ AC(post-training) & \underline{0.3827} & 0.7160 & \underline{0.5280} \\
%       Wan2.1 w/ FSG(post-training) & 0.3804 & \underline{0.7280} & 0.5240 \\
%       FG+Wan2.1 w/ post-training & \textbf{0.3860} & \textbf{0.7500} & \textbf{0.5320} \\
%       \bottomrule
%     \end{tabular}%
%   }
%   \caption{Ablation study results. Best scores are in \textbf{bold} and second best are \underline{underlined}. FG achieves significant performance gains with minimal post-training.}
%   \label{tab:Ablation}
%   \vspace{-4mm}
% \end{table}

\begin{table}[h!]
  \centering
  \resizebox{0.45\textwidth}{!}{%
    \begin{tabular}{lccc}
      \toprule
      \textbf{Method} & \makecell{\textbf{Dynamic}\\\textbf{Attributes}} & \makecell{\textbf{Human}\\\textbf{Motion}} & \makecell{\textbf{Human}\\\textbf{Interaction}} \\
      \midrule
      Wan2.1 & 0.3512 & 0.6920 & 0.4880 \\
      Wan2.1 w/ post-training & 0.3628 & 0.6980 & 0.5140 \\
      FG+Wan2.1 w/ zero-shot & 0.3512 & 0.7020 & 0.5220 \\
      Wan2.1 w/ AC(post-training) & \underline{0.3827} & 0.7160 & \underline{0.5280} \\
      Wan2.1 w/ FSG(post-training) & 0.3804 & \underline{0.7280} & 0.5240 \\
      FG+Wan2.1 w/ post-training & \textbf{0.3860} & \textbf{0.7500} & \textbf{0.5320} \\
      \bottomrule
    \end{tabular}%
  }
  \caption{Ablation study results on Wan2.1-I2V. Best scores are in \textbf{bold} and second best are \underline{underlined}. FG achieves significant performance gains with minimal post-training.}
  \label{tab:Ablation}
  \vspace{-4mm}
\end{table}

\subsection{Ablation Study}
We conduct a comprehensive ablation study to disentangle the contributions of FG. We evaluate its impact on \emph{Dynamic Attributes}, \emph{Human Motion}, and \emph{Human Interaction} against the Wan2.1-I2V baseline, with results summarized in \Cref{tab:Ablation}, leading to the following conclusions:
\begin{itemize}
    \item \textbf{Limited Gains from Standard Post-training.} Post-training with a small, general-purpose dataset yields marginal improvements (e.g., \emph{Human Motion}: 0.6920 $\to$ 0.6980), establishing a baseline but showing limited effectiveness due to the scale and general nature of the data.
    \item \textbf{FG Delivers Strong Gains without Fine-tuning.} Our FG module, without any post-training, significantly boosts performance on \emph{Human Motion} (0.7020) and \emph{Human Interaction} (0.5220), demonstrating its strong intrinsic capability to enhance semantic control.
    \item \textbf{FG and Post-training are Synergistic.} The full model combining both FG and post-training achieves the best performance, confirming the complementarity of explicit guidance and data-driven learning.
\end{itemize}
\vspace{-2mm}

\section{Conclusion, Limitation and Future Work}
We analyze the DiT-based I2V models and identify a key issue: while most layers respond well to semantic instructions, certain layers called Semantic-Weak Layers are less sensitive to text prompts. This weak responsiveness limits the model’s ability to generate videos aligned with text, causing over-reliance on visual priors. To address this, we propose Focal  Guidance which correct this issue and improves text controllability. We also design a benchmark to automatically assess how well videos align with their corresponding prompts. While FSG’s effectiveness is influenced by the underlying image encoder and the model's base capabilities. If the base model is weak, FSG will be constrained, as it relies on accurate semantic injection through the Attention Cache and the model’s fundamental capabilities.

\newpage
{
    \small
    \bibliographystyle{ieeenat_fullname}
    \bibliography{main}
}

\newpage
\appendix
\crefalias{section}{appendix}
\crefalias{subsection}{appendix}
\appendix
\onecolumn
\renewcommand\thesubsection{\Alph{subsection}} 
\renewcommand\thefigure{\Alph{subsection}.\arabic{figure}} 
\renewcommand\thetable{\Alph{subsection}.\arabic{table}} 
\setcounter{figure}{0} 
\setcounter{table}{0}
\section*{Appendix}

\begin{figure*}[htbp]
    \centering
    \includegraphics[width=0.99\textwidth]{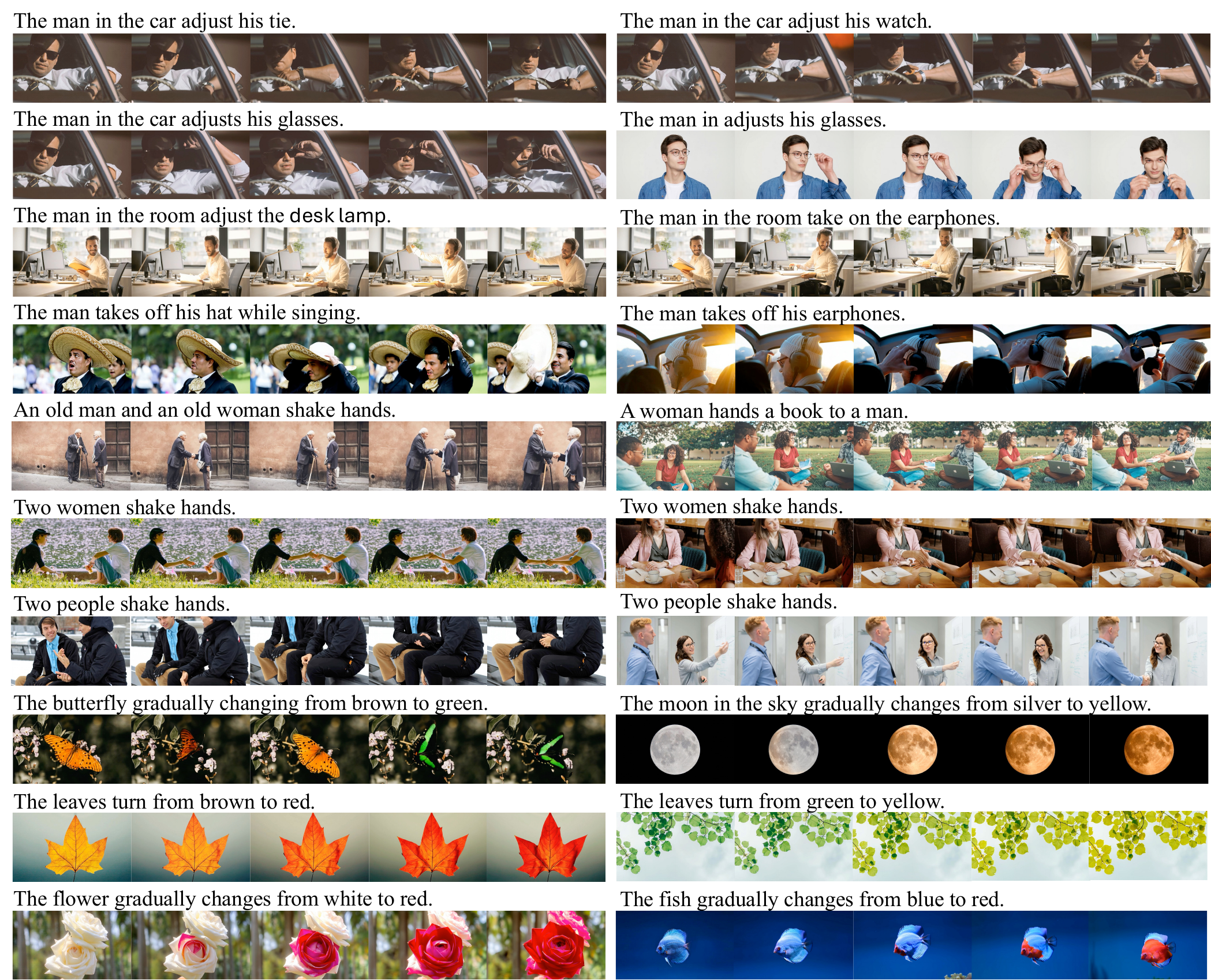}
    \caption{Illustrative qualitative examples generated by FG-Wan2.1-I2V 14B across three dimensions: human motion, human interaction, and dynamic attribute changes. These cases demonstrate the model’s ability to produce realistic, temporally consistent, and semantically coherent video outputs under diverse scenarios.}
    \label{fig:appendix_case}
    \vspace{-4mm}
\end{figure*}
\subsection{Experimental Setup}\label{appendix:exp}

\paragraph{Dataset} As an efficient method to unlock controllability in I2V models, FG aims to enhance model generation control with minimal post-training on limited data. We utilize an internally curated dataset of 12K samples with accurate captions with accurate captions, generated using Qwen2.5-VL-32B~\cite{bai2025qwen2}. The training objective is to teach the model this conditioning injection paradigm while preserving its original capabilities. This approach is model-agnostic, meaning it is applicable regardless of the underlying model or dataset.

\paragraph{Implementation Details.}
We evaluate FG on the CrossDiT-based Wan2.1-I2V~\citep{wan2025wan} and the MMDiT-based HunyuanVideo-I2V~\citep{kong2024hunyuanvideo}. 
For Wan2.1-I2V, we adopt the Wan I2V-14B-480P configuration and fine-tune the cross-attention layers in the Semantic-Weak Layers (layers 11--26) using a batch size of 8 and a learning rate of 1e-5, while applying FG throughout. 
For HunyuanVideo-I2V, we inject visual anchors via a CLIP encoder and apply FG to the single-stream layers 17--32, training with a batch size of 8 and a learning rate of 1e-4.

\begin{figure*}[htbp]
    \centering
    \includegraphics[width=0.99\textwidth]{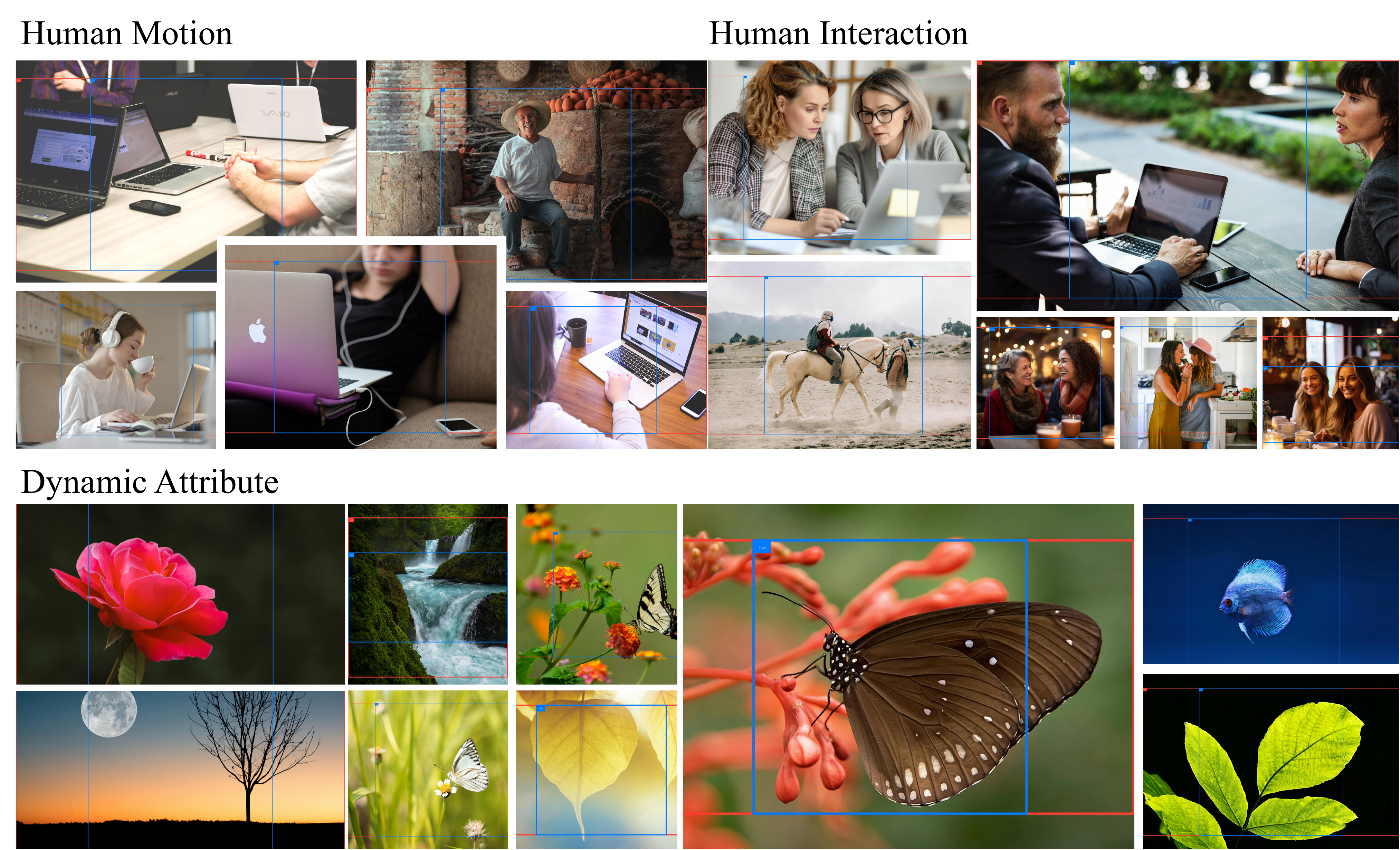}
    \caption{Visualization of reference images in our benchmark. We manually annotate subject bounding boxes on the original-resolution images and derive two canonical crops \textbf{16:9} and \textbf{1:1}. Based on these annotations we can generate adaptive resolution reference images for image to video generation.}
    \label{fig:bench}
    \vspace{-4mm}
\end{figure*}

\subsection{Controllability Evaluation and Dataset Annotation}\label{appendix:metric}

\paragraph{Metric Design.}
Current evaluation metrics for Image-to-Video (I2V) generation primarily focus on visual quality and subject consistency~\cite{zheng2025vbench,sauer2021projected,salimans2016improved,unterthiner2018towards,heusel2017gans,radford2021learning}, with limited attention to controllability. To fill this gap, we introduce three evaluation dimensions targeting instruction following: \textbf{dynamic attributes}, \textbf{human motion}, and \textbf{human interaction}. These dimensions enable a comprehensive assessment of how well generated videos adhere to both textual and visual conditions, promoting semantically grounded alignment between the text prompt and the reference image.
\begin{wrapfigure}{r}{0.55\textwidth}
    \vspace{-8pt}
    \centering
    \includegraphics[width=\linewidth]{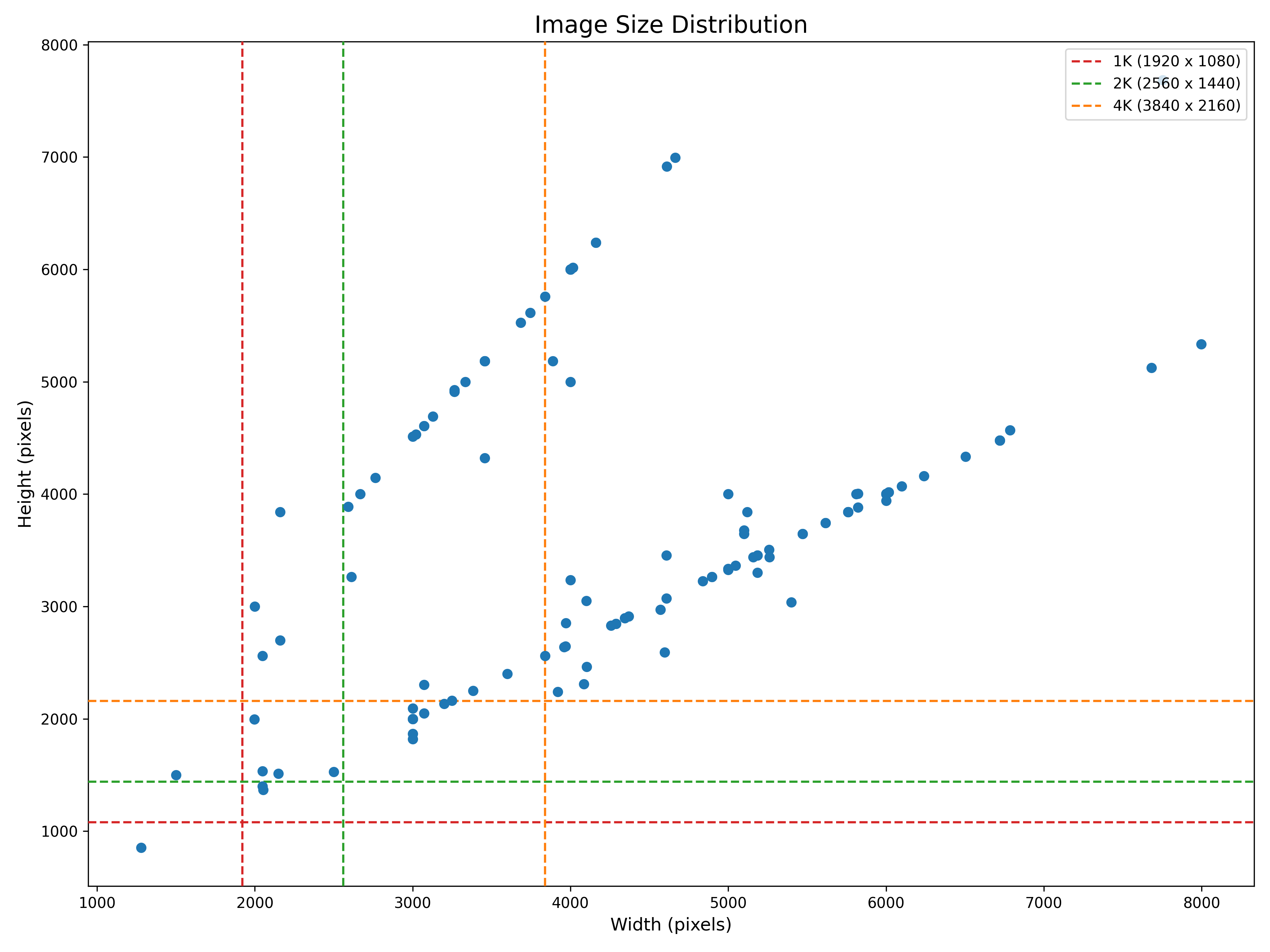}
    \caption{Resolution statistics of reference images.}
    \label{fig:image_size_distribution}
    \vspace{-6mm}
\end{wrapfigure}
To accurately evaluate whether the actions or attributes in the first-frame image are faithfully generated according to the text, we adopt a video-based multi-question answering (VQA) framework~\cite{zheng2025vbench}. Constrained by the first-frame reference, I2V has lower content freedom than T2V, and this framework mitigates evaluation noise while ensuring consistency across prompts. For each prompt, we design multiple complementary (and occasionally slightly redundant) questions to robustly check instruction following:
\begin{equation}
\text{Answer} = \sum_{i=1}^{N} \text{VQA}(Q_i, V \mid S),
\end{equation}
where $Q$ is the set of questions, $V$ is the video, and $S$ denotes the semantic structure of the prompt. The evaluation score is determined by whether all answers are correct.
\paragraph{Dataset Annotation and Cropping.}
For the three evaluation dimensions—Dynamic Attributes, Human Motion, and Human Interaction—we manually annotated datasets comprising 86, 100, and 100 image–prompt pairs, respectively, yielding 258, 278, and 303 corresponding questions. Following VBench-I2V~\cite{zheng2025vbench}, we annotate each image with the subject's bounding box and apply an aspect-ratio–aware cropping protocol to ensure the subject remains visible in all crops. Specifically:
(i) for portrait images $(\text{height} > \text{width})$, we first apply a 16:9 crop and then a 1:1 crop;
(ii) for landscape images $(\text{width} > \text{height})$, we first apply a 1:1 crop and then a 16:9 crop.
We maintain the original image resolutions; their distribution is shown in \Cref{fig:image_size_distribution}.

\subsection{Qualitative Results}\label{appendix:qualitative}
We present additional qualitative results on the best-performing model, FG-Wan2.1-I2V 14B, along three dimensions: human motion, human interaction, and dynamic attributes (as shown in~\Cref{fig:appendix_case}. These examples further illustrate the model’s strengths and its ability to generate realistic, coherent, and temporally expressive videos under various scenarios.

\end{document}